%% file: emnlp-multimodal.tex
\documentclass[11pt,a4paper]{article}
\usepackage[dvipsnames,usenames,svgnames]{xcolor}
\usepackage[hyperref]{emnlp2018}
\usepackage{times}
\usepackage{latexsym}
\usepackage{url}
\usepackage{graphicx} 
\usepackage[utf8]{inputenc} 
\usepackage[T1]{fontenc}    
\usepackage{url}            
\usepackage{booktabs}       
\usepackage{amsfonts}       
\usepackage{nicefrac}       
\usepackage{microtype}      
\usepackage[inline]{enumitem}
\usepackage{algorithm}
\usepackage{algorithmic}
\usepackage{float}
\usepackage{tikz}

\usepackage{latexsym}
\usepackage{amsmath}
\usepackage{amsthm}
\usepackage{amssymb}
\usepackage{mathtools}
\usepackage{multirow}
\usepackage{bm}
\usepackage{soul}
\usepackage{lscape}
\usepackage{graphicx,epstopdf}
\usepackage{psfrag}
\usepackage{rotating}
\usepackage{outlines}
\usepackage[textsize=small]{todonotes}
\usepackage{subcaption}
\usepackage{textcomp}
\usepackage{wrapfig}

\newcommand{\postspace}{\vskip -3mm}
\newcommand{\relLabel}[1]{\texttt{#1}}

\aclfinalcopy 

\newcommand{\para}[1]{\vskip 1mm\noindent\textbf{#1}~}

\input{macros}

\title{Embedding Multimodal Relational Data for Knowledge Base Completion}

\author{
  Pouya Pezeshkpour\\
  University of California\\
  Irvine, CA \\
  \texttt{pezeshkp@uci.edu}
  \And
  Liyan Chen \\
  University of California\\
  Irvine, CA \\
  \texttt{liyanc@uci.edu} \\
  \And
  Sameer Singh \\
  University of California\\
  Irvine, CA \\
  \texttt{sameer@uci.edu} \\
}

\date{}

\begin{document}
\maketitle

\begin{abstract}

Representing entities and relations in an embedding space is a well-studied approach for machine learning on relational data. Existing approaches, however, primarily focus on simple link structure between a finite set of entities, ignoring the variety of data types that are often used in knowledge bases, such as text, images, and numerical values. In this paper, we propose multimodal knowledge base embeddings (MKBE) that use different neural encoders for this variety of observed data, and combine them with existing relational models to learn embeddings of the entities and multimodal data. Further, using these learned embedings and different neural \emph{decoders}, we introduce a novel multimodal imputation model to generate missing multimodal values, like text and images, from information in the knowledge base. We enrich existing relational datasets to create two novel benchmarks that contain additional information such as textual descriptions and images of the original entities. We demonstrate that our models utilize this additional information effectively to provide more accurate link prediction, achieving state-of-the-art results with a considerable gap of 5-7$\%$ over existing methods. Further, we evaluate the quality of our generated multimodal values via a user study. We have release the datasets and the open-source implementation of our models at \url{https://github.com/pouyapez/mkbe}. 
\end{abstract}


\section{Introduction}
\label{sec:intro}

Knowledge bases (KB) are an essential part of many computational systems with applications in search,  structured data management, recommendations, question answering, and information retrieval. 
However, KBs often suffer from incompleteness, noise in their entries, and inefficient inference under uncertainty.
To address these issues, learning relational knowledge representations has been a focus of active research~\citep{bordes2011learning,bordes2013translating,yang2014embedding,nickel2016holographic,trouillon2016complex,dettmers2017convolutional}. 
These approaches represent relational triples, that consist of a subject entity, relation, and an object entity, by learning fixed, low-dimensional representations for each entity and relation from observations, encoding the uncertainty and inferring missing facts accurately and efficiently. 
The subject and the object entities come from a fixed, enumerable set of entities that appear in the knowledge base. 

\begin{figure}[tb] 
  \begin{center}
    \includegraphics[width=0.45\textwidth]{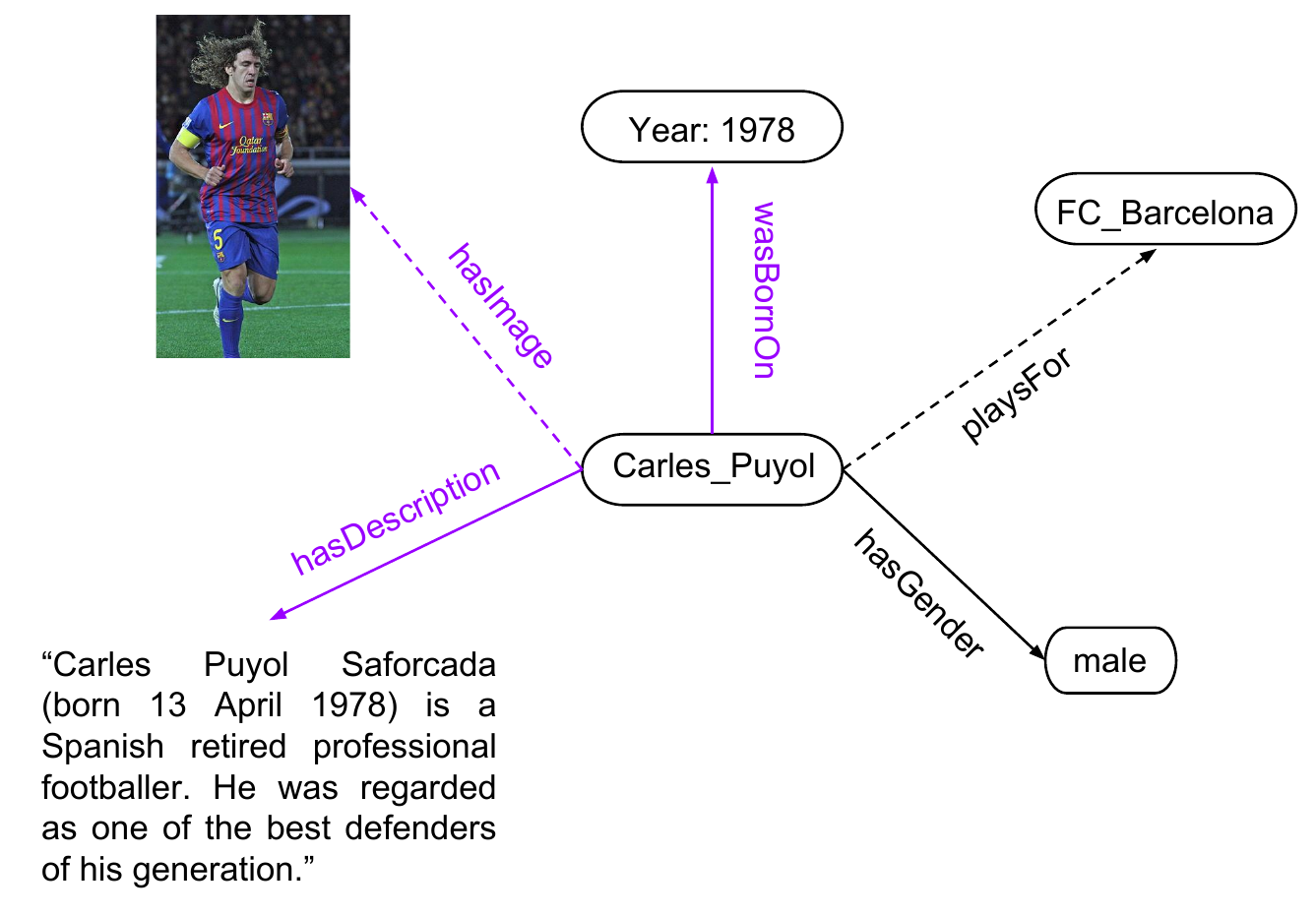}
  \end{center}
  \postspace{}
  \caption{{\bf Example of a Multimodal KB.} Graph representation of (a part of) a KB that consists of regular links (in black) and multimodal ones (in \textcolor{violet}{purple}) that we support in this work. } 
  \label{fig:graph}
    \postspace{}
\end{figure}

Knowledge bases in the real world, however, contain a wide variety of data types beyond these direct links.
Apart from relations to a fixed set of entities, KBs often not only include numerical attributes (such as ages, dates, financial, and geoinformation), but also textual attributes (such as names, descriptions, and titles/designations) and images~(profile photos, flags, posters, etc.).
These different types of data can play a crucial role as extra pieces of evidence for knowledge base completion. 
For example the textual descriptions and images might provide evidence for a person's age, profession, and designation. 
In the multimodal KB shown in Figure~\ref{fig:graph} for example, the image can be helpful in predicting of Carles Puyol's occupation, while the description contains his nationality. 
Incorporating this information into existing approaches as entities, unfortunately, is challenging as they assign each entity a distinct vector and predict missing links (or attributes) by enumerating over the possible values, both of which are only possible if the entities come from a small, enumerable set.
There is thus a crucial need for relational modeling that goes beyond just the link-based view of KB completion, by not only utilizing multimodal information for better link prediction between existing entities, but also being able to generate missing multimodal values. 

In this paper, we introduce \emph{multimodal knowledge base embeddings}~(MKBE) for modeling knowledge bases that contain a variety of data types, such as links, text, images, numerical, and categorical values.
We propose neural encoders and decoders to replace initial layers of any embedding-based relational model; we apply them to DistMult~\citep{yang2014embedding} and ConvE~\citep{dettmers2017convolutional} here. 
Specifically, instead of learning a distinct vector for each entity and using enumeration to predict links, MKBE includes the following extensions: (1)~introduce additional neural encoders to embed multimodal evidence types that the relational model uses to predict links, and (2)~introduce neural decoders that use an entity's embedding to generate its multimodal attributes (like image and text).
For example, when the object of a triple is an image, we encode it into a fixed-length vector using a CNN, while textual objects are encoded using RNN-based sequence encoders. 
The scoring module remains identical to the underlying relational model; given the vector representations of the subject, relation, and object of a triple, we produce a score indicating the probability that the triple is correct using DistMult or ConvE.
After learning the KB representation, neural decoders use entity embeddings to generate missing multimodal attributes, for example, generating the description of a person from their structured information in the KB. 
This unified framework allows for flow of the information across the different relation types (multimodal or otherwise), providing a more accurate modeling of relational data. 

We provide an evaluation of our proposed approach on two relational KBs.
Since we are introducing the multimodal KB completion setting, we provide two benchmarks, created by extending the existing YAGO-10 and MovieLens-100k datasets to include additional relations such as textual descriptions, numerical attributes, and images of the entities.
We demonstrate that MKBE utilizes the additional information effectively to provide gains in link-prediction accuracy, achieving state-of-the-art results on these datasets for both the DistMult and the ConvE scoring functions. 
We evaluate the quality of multimodal attributes generated by the decoders via user studies that demonstrate their realism and information content, along with presenting examples of such generated text and images.


\section{Multimodal KB Completion} 
As described earlier, KBs often contain different types of information about entities including links, textual descriptions, categorical attributes, numerical values, and images.
In this section, we briefly introduce existing relational embedding approaches that focus on modeling the linked data using distinct, dense vectors.
We then describe MKBE that extends these approaches to the multimodal setting, i.e., modeling the KB using all the different information to predict the missing links and impute the missing attributes.

\subsection{Background on Link Prediction}
Factual statements in a knowledge base are represented using a triple of subject, relation, and object, $\langle s, r, o\rangle$, where $s,o\in\xi$, a set of entities, and $r\in \R$, a set of relations.
Respectively, we consider two goals for relational modeling, (1)~to train a machine learning model that can score the \emph{truth} value of any factual statement, and (2)~to predict missing links between the entities. 
In existing approaches, a scoring function $\psi:\xi\times \R\times\xi\rightarrow\Real$ (or sometimes, $[0,1]$) is learned to evaluate whether any given fact is true, as per the model. 
For predicting links between the entities, since the set $\xi$ is small enough to be enumerated, missing links of the form $\langle s,r,?\rangle$ are identified by enumerating all the objects and scoring the triples using $\psi$ (i.e. assume the resulting entity comes from a known set). 
For example, in Figure~\ref{fig:graph}, the goal is to predict that Carles Puyol plays for Barcelona.

Many of the recent advances in link prediction use an embedding-based approach; each entity in $\xi$ and relation in $\R$ are assigned distinct, dense vectors, which are then used by $\psi$ to compute the score. 
In DistMult~\citep{yang2014embedding}, for example, each entity~$i$ is mapped to a $d$-dimensional dense vector ($\mathbf{e}_i\in\Real^{d}$) and each relation $r$ to a diagonal matrix $\mathbf{R}_r\in\Real^{d\times d}$, and consequently, the score for any triple $\langle s,r,o\rangle$ is computed as $\psi(s,r,o) = \mathbf{e}_s^T \mathbf{R}_r \mathbf{e}_o$. 
Along similar lines, ConvE~\citep{dettmers2017convolutional} uses vectors to represent the entities and the relations, $\mathbf{e}_s,\mathbf{e}_o,\mathbf{r}_r\in\Real^{d\times 1}$, then, after applying a CNN layer on $\mathbf{e}_s$ and $\mathbf{r}_r$, combines it with $\mathbf{e}_o$ to score a triplet, i.e. the scoring function $\psi(s,r,o)$ is $f(\text{vec}(f([\bar{\mathbf{e}_s};\bar{\mathbf{r}_r}*w]))W)\mathbf{e}_o$. 
Other relational embedding approaches primarily vary in their design of the scoring function~\cite{bordes2013translating,yang2014embedding,nickel2016holographic,trouillon2016complex}, but share the shortcoming of assigning distinct vectors to every entity, and assuming that the possible object entities can be enumerated.
In this work we focus on DistMult because of its simplicity, popularity, and high accuracy, and ConvE because of its state-of-the-art results.

\subsection{Problem Setup}

When faced with additional triples in form of multimodal data, the setup of link prediction is slightly different.
Consider a set of all potential multimodal objects, $\M$, i.e. possible images, text, numerical, and categorical values, and multimodal evidence triples, $\langle s,r,o\rangle$, where $s\in\xi$, $r\in\R$, and $o\in\M$.
Our goals with incorporating multimodal information into KB remain the same: we want to be able to score the truth of any triple $\langle s,r,o\rangle$, where $o$ is from $\xi$ (link data) or from $\M$ (multimodal data), and to be able to predict missing value $\langle s,r,?\rangle$ that may be from $\xi$ or $\M$ (depending on $r$).
For the example in Figure~\ref{fig:graph}, in addition to predicting that Carles Puyol plays for Barcelona from multimodal evidence, we are also interested in generating an image for Carles Puyol, if it is missing.

Existing approaches to this problem assume that the subjects and the objects are from a fixed set of entities $\xi$, and thus are treated as indices into that set, which fails for the multimodal setting primarily for two reasons.
First, learning distinct vectors for each object entity does not apply to multimodal values as they will ignore the actual content of the multimodal attribute.
For example, there will be no way to generalize vectors learned during training to unseen values that might appear in the test; this is not a problem for the standard setup due to the assumption that all entities have been observed during training.
Second, in order to predict a missing \emph{multimodal} value, $\langle s,r,?\rangle$, enumeration is not possible as the search space is potentially infinite (or at least intractable to search).

\begin{figure*}[tb]
 \centering
    \includegraphics[width=0.95\textwidth]{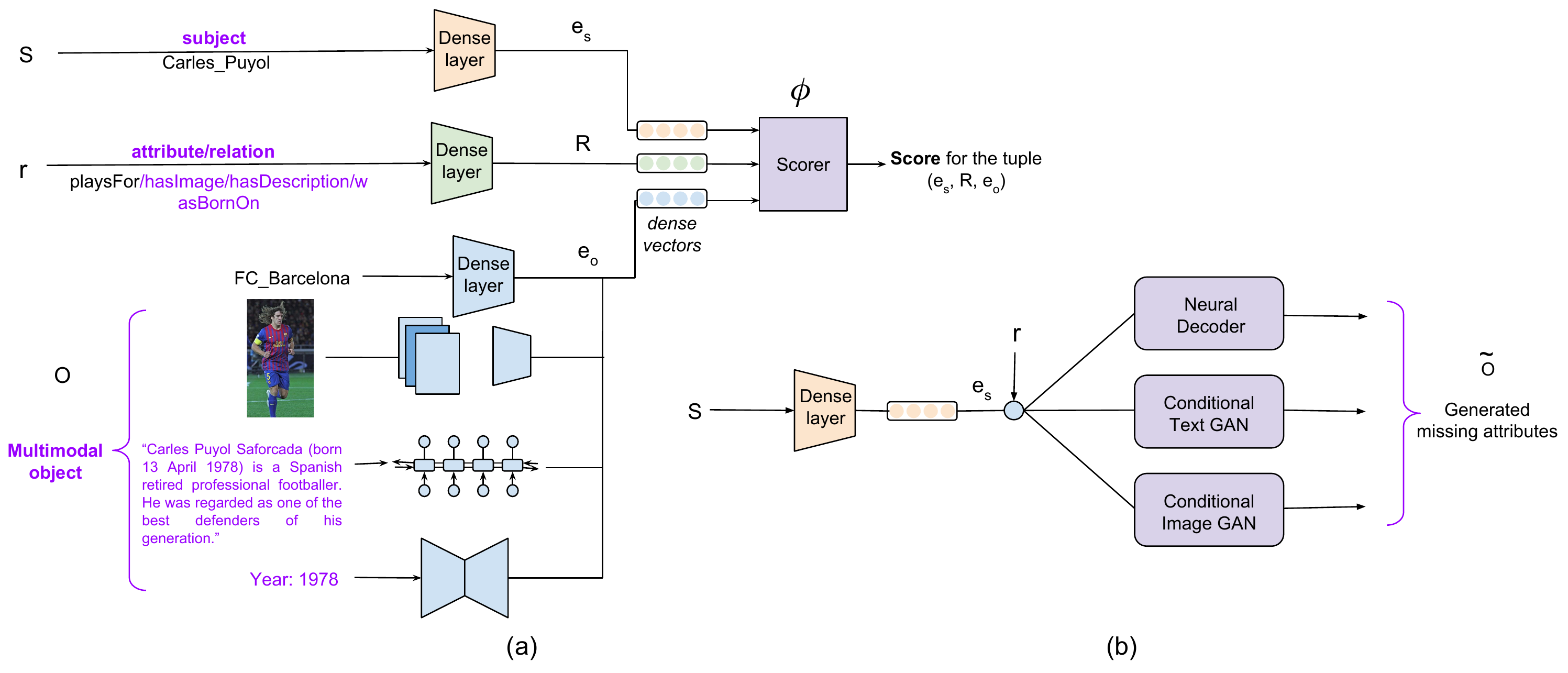}
\postspace{}
  \caption{\textbf{Multimodal KB Embeddings (MKBE):} (a) Proposed architecture that, given any entity and its relations
  , uses domain-specific encoders to embed each object. The embeddings of entities, and the relation are then used to score the \emph{truth} value of the triple by the \emph{Scorer}.
  (b) Architecture of the proposed work for multimodal attributes recovery. Given an entity, we use its learned embeddings from (a) as the context for attribute-specific decoders to generate the missing values.}
  \label{fig:overall-net}
    \postspace{}
\end{figure*}

\subsection{Multimodal KB Embeddings (MKBE)}
To incorporate such multimodal \emph{objects} into the existing relational models like DistMult and ConvE, we propose to learn embeddings for these types of data as well.
We utilize recent advances in deep learning to construct \emph{encoders} for these objects to represent them, essentially providing an embedding $\mathbf{e}_o$ for any object value.

The overall goal remains the same: the model needs to utilize all the observed subjects, objects, and relations, across different data types, in order to estimate whether any fact $\langle s, r, o\rangle$ holds. 
We present an example of an instantiation of MKBE for a knowledge base containing YAGO entities in Figure~\ref{fig:overall-net}a.
For any triple $\langle s,r,o\rangle$, we embed the subject (\emph{Carles Puyol}) and the relation (such as \relLabel{playsFor}, \relLabel{wasBornOn}, or \relLabel{playsFor}) using a direct lookup.
For the object, depending on the domain (indexed, string, numerical, or image, respectively), we use approrpiate encoders to compute its embedding $\mathbf{e}_o$.
As in DistMult and ConvE, these embeddings are used to compute the score of the triple.

Via these neural encoders, the model can use the information content of multimodal objects to predict missing links where the objects are from $\xi$, however, learning embeddings for objects in $\M$ is not sufficient to generate missing multimodal values, i.e. $\langle s, r, ?\rangle$ where the object is in $\M$.
Consequently, we introduce a set of neural decoders $D:\xi\times \R\rightarrow\M$ that use entity embeddings to \emph{generate} multimodal values. An outline of our model for imputing missing values is depicted in Figure~\ref{fig:overall-net}b. 
We will describe these decoders in Section~\ref{sec:decoders}.

\subsection{Encoding Multimodal Data}
\label{sec:encoders}

Here we describe the encoders we use for multimodal objects. A simple example of MKBE is provided in Figure~\ref{fig:overall-net}a. As it shows, we use different encoder to embed each specific data type. 

\para{Structured Knowledge}
 Consider a triplet of information in the form of $\langle s ,r ,o\rangle$. 
 To represent the subject entity $s$ and the relation $r$ as independent embedding vectors (as in previous work), we pass their one-hot encoding through a dense layer. 
 Furthermore, for the case that the object entity is categorical, we embed it through a dense layer with a recently introduced selu activation~\citep{klambauer2017self}, with the same number of nodes as the embedding space dimension. 

\para{Numerical}
Objects in the form of real numbers can provide a useful source of information and are often quite readily available. 
We use a feed forward layer, after standardizing the input, in order to \emph{embed} the numbers (in fact, we are projecting them to a higher-dimensional space, from $\Real\rightarrow\Real^d$). 
It is worth noting that existing methods treat numbers as distinct \emph{entities}, e.g., learn independent vectors for numbers \texttt{39} and \texttt{40}, relying on data to learn that these values are similar to each other. 
 
 
\para{Text}
Since text can be used to store a wide variety of different types of information, for example names versus paragraph-long descriptions, we create different encoders depending on the lengths of the strings involved.
For attributes that are fairly short, such as names and titles, we use character-based stacked, bidirectional GRUs to encode them, similar to~\citet{verga2015multilingual}, using the final output of the top layer as the representation of the string.
For strings that are much longer, such as detailed descriptions of entities consisting of multiple sentences, we treat them as a sequence of words, and use a CNN over the word embeddings, similar to ~\citet{francis2016capturing}, in order to learn the embedding of such values.
These two encoders provide a fixed length encoding that has been shown to be an accurate semantic representation of strings for multiple tasks
~\citep{dos2014deep}.

 
 
\para{Images}
Images can also provide useful evidence for modeling entities. 
For example, we can extract person's details such as gender, age, job, etc., from image of the person~\citep{levi2015age}, or location information such as its approximate coordinates, neighboring locations, and size from map images~\citep{weyand2016planet}.
A variety of models have been used to compactly represent the semantic information in the images, and have been successfully applied to tasks such as image classification, captioning~\citep{karpathy2015deep}, and question-answering~\citep{yang2016stacked}.
To embed images such that the encoding represents such semantic information, we use the last hidden layer of VGG pretrained network on Imagenet~\citep{simonyan2014very}, followed by compact bilinear pooling~\citep{gao2016compact}, to obtain the embedding of the images.


\para{Training}
We follow the setup from \citet{dettmers2017convolutional} that consists of binary cross-entropy loss without negative sampling for both ConvE and DisMult scoring.
In particular, for a given subject-relation pair $(s,r)$, we use a binary \emph{label} vector $\mathbf{t}^{s,r}$ over all entities, indicating whether $\langle s,r,o\rangle$ is observed during training.
Further, we denote the model's probability of truth for any triple $\langle s,r,o\rangle$ by $p^{s,r}_o$, computed using a sigmoid over $\psi(s,r,o)$.
The binary cross-entropy loss is thus defined as:
\begin{align*}
    \nonumber
    \sum_{(s,r)}\sum_{o} t^{s,r}_o\log(p^{s,r}_o) + (1-t^{s,r}_o)\log(1 - p^{s,r}_o).
\end{align*}
We use the same loss for multimodal triples as well, except that the summation is restricted to the objects of the same modality, i.e. for an entity $s$ and its text description, $\mathbf{t}^{s,r}$ is a one-hot vector over all descriptions observed during training.

\subsection{Decoding Multimodal Data}
\label{sec:decoders}

Here we describe the decoders we use to generate multimodal values for entities from their embeddings. 
The multimodal imputing model is shown in Figure~\ref{fig:overall-net}b, which uses different neural decoders to generate missing attributes (more details are provided in supplementary materials). 

\para{Numerical and Categorical data}
To recover the missing numerical and categorical data such as dates, gender, and occupation, we use a simple feed-forward network on the entity embedding to predict the missing attributes.
In other words, we are asking the model, if the actual birth date of an entity is not in the KB, what will be the most likely date, given the rest of the relational information. 
These decoders are trained with embeddings from Section~\ref{sec:encoders}, with  appropriate losses (RMSE for numerical and cross-entropy for categories).  

\para{Text}
A number of methods have considered generative adversarial networks (GANs) to generate grammatical and linguistically coherent sentences~\citep{yu2017seqgan,rajeswar2017adversarial,guo2017long}. 
In this work, we use the adversarially regularized autoencoder (ARAE)~\citep{kim2017adversarially} to train generators that decodes text from continuous codes, however, instead of using the random noise vector $z$, we condition the generator on the entity embeddings. 
\para{Images}
Similar to text recovery, to find the missing images we use conditional GAN structure. Specifically, we combine the BE-GAN~\citep{berthelot2017began} structure with pix2pix-GAN~\citep{isola2017image} model to generate high-quality images, conditioning the generator on the entity embeddings in the knowledge base representation. 


\section{Related Work}
\label{sec:related}
There is a rich literature on modeling knowledge bases using low-dimensional representations, differing in the operator used to score the triples. 
In particular, they use matrix and tensor multiplication ~\citep{nickel2011three,yang2014embedding,socher2013reasoning}, Euclidean distance~\citep{bordes2013translating,wang2014knowledge,lin2015learning}, circular correlation ~\citep{nickel2016holographic}, or the Hermitian dot product~\citep{trouillon2016complex} as scoring function.
However, the \emph{objects} for all of these approaches are a fixed set of entities, i.e., they only embed the structured links between the entities. 
Here, we use different types of information (text, numerical values, images, etc.) in the encoding component by treating them as relational triples. 


A number of methods utilize an extra type of information as the observed features for entities, by either merging, concatenating, or averaging the entity and its features to compute its embeddings, such as numerical values~\citep{garcia2017kblrn} (we use KBLN from this work to compare it with our approach using only numerical as extra attributes), images~\citep{xie2016image,onoro2017representation} (we use IKRL from the first work to compare it with our approach using only images as extra attributes), text~\citep{mcauley2013hidden,zhong2015aligning,toutanova2015representing,toutanova2016compositional,xie2016representation,tu2017cane}, and a combination of text and image~\citep{sergieh2018multimodal}. Further, \citet{verga2015multilingual} address the multilingual relation extraction task to attain a universal schema by considering raw text with no annotation as extra feature and using matrix factorization to jointly embed KB and textual relations~\citep{riedel13:relation}.
In addition to treating the extra information as features, graph embedding approaches~\citep{schlichtkrull2017modeling,kipf2016variational} consider observed attributes while encoding to achieve more accurate embeddings.

The difference between MKBE and these mentioned approaches is three-fold: (1)~we are the first to use different types of information in a unified model, (2)~we treat these different types of information~(numerical, text, image) as relational triples of structured knowledge instead of predetermined features, i.e., first-class citizens of the KB, and not auxiliary features, and (3)~our model represents uncertainty in them, supporting the missing values and facilitating recovery of missing values. 

\begin{table}[tb]
  \centering
        \caption{\textbf{Data Statistics} of the two benchmark datasets we are using. The numbers in bold are our contributions to the datasets.}
        \label{tab:data_stats}
        \small
        \begin{tabular}{lrr}
        \toprule
        & \bf MovieLens & \bf YAGO-10 \\
        \midrule
        \#Link Types & 13 &  45\\
        \#Entities & 2,625 & 123,182 \\
        \#Link Triples & 100,000 & 1,079,040 \\\addlinespace[3pt]
        \#Numerical Attributes & 2,625 & \bf 111,406 \\
        \#Image Attributes & \bf 1,651 & \bf 61,246\\
        \#Text Attributes & 1,682 &  \bf 107,326\\
        \bottomrule
        \end{tabular}
        \postspace{}
\end{table}

\section{Evaluation Benchmarks}

To evaluate the performance of our multimodal relational embeddings approach, we provide two new benchmarks by extending existing datasets. 
Table~\ref{tab:data_stats} provides the statistics of these datasets. 

\para{MovieLens-100k}
dataset~\citep{harper2016MovieLens} 
is a popular benchmark in recommendation systems to predict user ratings with contextual features, containing around $1000$ users on $1700$ movies. 
MovieLens already contains rich relational data about occupation, gender, zip code, and age for users and genre, release date, and the titles for movies. 
We augment this data with movie posters collected from TMDB (\url{https://www.themoviedb.org/}). 
We treat the 5-point ratings as five different relations in KB triple format, i.e., $\langle\text{user},r=5,\text{movie}\rangle$, and evaluate the rating predictions as other relations are introduced. 

\para{YAGO-10}
Even though MovieLens has a variety of data types, it is still quite small, and is over a specialized domain. 
We also consider a second dataset that is much more appropriate for knowledge graph completion and is popular for link prediction, the YAGO3-10 knowledge graph ~\citep{suchanek2007yago,nickel2012factorizing}.  
This graph consists of around 120,000 entities, such as people, locations, and organizations, and 37 relations, such as kinship, employment, and residency, and thus much closer to the traditional information extraction goals. 
We extend this dataset with the textual description (as an additional relation) and the images associated with each entity (for half of the entities), provided by \emph{DBpedia}~\citep{lehmann2015dbpedia}. 
We also include additional relations such as \relLabel{wasBornOnDate} that have dates as values.

    \begin{table}[tb]
    \centering
    \small
        \caption{\textbf{Rating Prediction in MovieLens.} Results for models that use: rating information (R), movie-attribute (M), user-attribute (U), movies' title text (T), and poster images (P).}
        \label{tab:r-ml}
 	        \setlength{\tabcolsep}{5pt}
       \begin{tabular}{clcccc}
        \toprule
        & \bf Models&\bf MRR& \bf Hits@1& \bf Hits@2& \bf RMSE\\ \midrule
        \parbox[t]{2mm}{\multirow{5}{*}{\rotatebox[origin=c]{90}{\bf DistMult}}} &
        Ratings Only, R&0.62&0.40&0.69&1.48\\ 
        & R+M+U&0.646&0.423&0.708&1.37\\
        & R+M+U+T&0.650&\bf{0.424}&\bf{0.73}&\bf{1.23}\\
        & R+M+U+P&\bf{0.652}&0.413&0.712&1.27\\
        & R+M+U+T+P&0.644&0.42&0.72&1.3\\
        \midrule
        \parbox[t]{2mm}{\multirow{5}{*}{\rotatebox[origin=c]{90}{\bf ConvE}}} &
        Ratings Only, R&0.683&0.47&0.81&1.47\\
        & R+M+U&0.702&0.49&0.83&1.39\\
        & R+M+U+T&\textbf{0.728}&\textbf{0.513}&\textbf{0.85}&1.13\\
        & R+M+U+P&0.726&0.512&0.83&1.13\\
        & R+M+U+T+P&0.726&0.512&0.84&\textbf{1.09}\\ 

        \bottomrule
    \end{tabular}
    \end{table}
\begin{table}[tb]
\small
	\centering
	\caption{\textbf{Link Prediction in YAGO-10.} Results shown for models using: structured information (S), textual description of the entities (D), dates as numerical information (N), and images (I). Published refers to \citet{dettmers2017convolutional}.}
	\label{tab:yago}
	        \setlength{\tabcolsep}{5pt}
\begin{tabular}{clcccc}
\toprule
& \bf Models & \bf MRR& \bf Hits@1& \bf Hits@3& \bf Hits@10\\ \midrule
\parbox[t]{2mm}{\multirow{7}{*}{\rotatebox[origin=c]{90}{\bf DistMult}}} &
Published
&0.337&0.237&0.379&0.54\\ 
& Links only, S &0.326&0.221&0.375&0.538\\ 
& S+D&0.36&0.262&0.395&0.571\\
& S+N&0.325&0.213&0.382&0.517\\
& S+I&0.342&0.235&0.352&0.618\\
& S+D+N&0.359&0.243&0.401&0.679\\
& S+D+N+I&\bf{0.372}&\bf{0.268}&\bf{0.418}&\bf{0.792}\\
\midrule
\parbox[t]{2mm}{\multirow{9}{*}{\rotatebox[origin=c]{90}{\bf ConvE}}} &
Published&0.523&0.448&0.564&0.658\\
& Links only, S&0.482&0.372&0.519&0.634\\
& S+D& 0.564&0.478&0.595&0.713\\
& S+N&0.549&0.462&0.587&0.701\\
& S+I&0.566&0.471&0.597&0.72\\
& S+D+N&\bf{0.588}&0.517&0.603&\bf{0.722}\\
& S+D+N+I&0.584&\bf{0.52}&\bf{0.604}&0.698\\
\cmidrule{2-6}
&KBLN&0.503&0.41&0.549&0.658\\
&IKRL&0.509&0.423&0.556&0.663\\
\bottomrule
\end{tabular}
\postspace{}
\end{table}

\section{Experiment Results}

In this section, we first evaluate the ability of MKBE to utilize the multimodal information by comparing to DistMult and ConvE through a variety of tasks. 
Then, by considering the recovery of missing multimodal values (text, images, and numerical) as the motivation, we examine the capability of our models in generation. 
Details of the hyperparameters and model configurations is provided in the supplementary material, and the source code and the datasets to reproduce the results is available at \url{https://github.com/pouyapez/mkbe}.
\begin{table*}[tb]
	\centering
	\small
	\caption{\textbf{Per-Relation Breakdown} showing performance of each model on different relations.}
	\label{tab:rel-yago}
\begin{tabular}{lcccccccc}
\toprule
\multirow{2}{*}{\bf Relation}&\multicolumn{2}{c}{\bf Links Only}&\multicolumn{2}{c}{\bf +Numbers}&\multicolumn{2}{c}{\bf +Description}&\multicolumn{2}{c}{\bf +Images}\\
\cmidrule(lr){2-3}
\cmidrule(lr){4-5}
\cmidrule(lr){6-7}
\cmidrule(lr){8-9}
 & MRR& Hits@1& MRR& Hits@1& MRR& Hits@1& MRR& Hits@1\\ \midrule
\relLabel{isAffiliatedTo}&0.524&0.401&0.551&0.467&\bf{0.572}&\bf{0.481}&0.569&0.478\\ 
\relLabel{playsFor}&0.528&0.413&0.554&0.471&\bf{0.574}&\bf{0.486}&0.566&0.476\\ 
\relLabel{hasGender}&0.798&0.596&0.799&0.599&0.813&0.627&\bf{0.842}&\bf{0.683}\\
\relLabel{isConnectedTo}&0.482&0.367&\bf{0.497}&0.379&0.492&\bf{0.384}&0.484&0.372\\
\relLabel{isMarriedTo}&0.365&0.207&0.387&0.221&0.404&0.296&\bf{0.413}&\bf{0.326}\\

\bottomrule
\end{tabular}

\end{table*}

\subsection{Link Prediction}

In this section, we evaluate the capability of MKBE in the link prediction task. 
The goal is to calculate MRR and Hits@ metric (ranking evaluations) of recovering the missing entities from triples in the test dataset, performed by ranking all the entities and computing the rank of the correct entity. Similar to previous work, here we focus on providing the results in a filtered setting, that is we only rank triples in the test data against the ones that never appear in either train or test datasets.

\para{MovieLens-100k}
We train the model using \relLabel{Rating} as the relation between users and movies. 
We use a character-level GRU for the movie titles
, a separate feed-forward network
for age, zip code, and release date, and finally, we use a VGG network on the posters (for every other relation we use a dense layer).
Table~\ref{tab:r-ml} shows the link (rating) prediction evaluation on MovieLens when test data is consisting only of rating triples.
We calculate our metrics by ranking the five relations that represent ratings instead of object entities. 
We label models that use ratings as R, movie-attributes as M, user-attributes as U, movie titles as T, and posters as P.
As shown, the model \emph{R+M+U+T} outperforms others with a considerable gap demonstrating the importance of incorporating extra information. 
Hits@1 for the baseline is \emph{40\%}, matching existing recommendation systems~\citep{guimera2012predicting}. 
From these results, we see that the models benefit more from titles as compared to the posters. 

\para{YAGO-10} The result of link prediction on our YAGO dataset is provided in Table~\ref{tab:yago}. 
We label models using structured information as S, entity-description as D, numerical information as N, and entity-image as I. 
We see that the model that encodes all type of information consistently performs better than other models, indicating that the model is effective in utilizing the extra information. 
On the other hand, the model that uses only text performs the second best, suggesting the entity descriptions contain more information than others. 
It is notable that model $S$ is outperformed by all other models, demonstrating the importance of using different data types for attaining higher accuracy. 
This observation is consistent across both DistMult and ConvE, and the results obtained on ConvE are the new state-of-art for this dataset (as compared to \citet{dettmers2017convolutional}). 
Furthermore, we implement KBLN~\citep{garcia2017kblrn} and  IKRL~\citep{xie2016image} to compare them with our S+N and S+I models. Our models outperform these approaches, in part because both of these methods require same multimodal attributes for both of the subject and object in each triple. 


\para{Relation Breakdown}
We perform additional analysis on the YAGO dataset to gain a deeper understanding of the performance of our model using ConvE method. Table~\ref{tab:rel-yago} compares our models on some of the most frequent relations. 
As shown, the model that includes textual description significantly benefits \texttt{isAffiliatedTo}, and \texttt{playsFor} relations, as this information often appears in text. 
Moreover, images are useful for \texttt{hasGender} and \texttt{isMarriedTo}, while for the relation \texttt{isConnectedTo}, numerical (dates) are more effective than images.  

\cut{
  \begin{table}[tb]
      \centering
        \caption{Predicting Genres in MovieLens}
        \label{fig:genre-ml}
            \small
            \begin{tabular}{lrrr}
            \toprule
            \bf Models& \bf MRR& \bf Hits@1&\bf Hist@10\\ \midrule
            R+M&0.074&0.014&0.175\\ 
            R+M+U&0.071&0.023&0.145\\ 
            R+M+U+T&0.075&0.020&0.163\\ 
            R+M+U+P&\bf{0.103}&0.038&0.223\\ 
            R+M+U+T+P&0.102&\bf{0.047}&\bf{0.232}\\ 
            \bottomrule
            \end{tabular}

    \end{table}
}
    \begin{table}[tb] 
     \caption{\textbf{Predicting Numbers and Categories} for YAGO (dates) and MovieLens (genres), using models with access with different information.}
      \centering
 \begin{subfigure}[b]{0.5\columnwidth} 
 \centering
        \small
        \setlength{\tabcolsep}{2pt}
            \begin{tabular}{lcc}
            \toprule
            {\bf Models} & \bf Search & \bf Decoding\\
            \midrule
            S+N &62.49&58.7\\
            S+N+D &59.42&56.2\\ 
            S+N+I  &59.86&55.8\\ 
            All Info  &\bf{57.62}&\bf{54.1}\\ 
            \bottomrule
            \end{tabular}
   \vspace{12pt}
  \caption{RMSE (years) in YAGO}
  \label{fig:ny}
  \end{subfigure}
 \begin{subfigure}[b]{0.45\columnwidth} 
 \centering
            \small
             \begin{tabular}{lc}
            \toprule
            \bf Models & \bf Accuracy\\
            \midrule
            R+M &71.82\\
            R+M+U &71.98\\ 
            R+M+U+T  & 73.01\\ 
            R+M+U+P  & 73.77\\ 
            All Info  & \bf{75.89}\\ 
            \bottomrule
            \end{tabular}
   \vspace{3pt}
  \caption{Genres in MovieLens}
  \label{fig:gm}
\end{subfigure}

                \postspace{}

  \end{table}

\subsection{Imputing Multimodal Attributes}
\label{sec:mr}

Here we present an evaluation on imputing multimodal attributes (text, image and numerical). 
\para{Numerical and Categorical}
Table~\ref{fig:ny} shows performance of predicting missing numerical attributes in the data, evaluated via holding out $10\%$ of the data. 
We only consider numerical values (dates) that are more recent than $1000 AD$ to focus on more relevant entities.
In addition to the neural decoder, we train a search-based decoder as well by considering all $1017$ choices in the interval $[1000,2017]$, and for each triple in the test data, finding the number that the model scores the highest; we use this value to compute the RMSE.
As we can see, \emph{all info} outperform other methods on both datasets, demonstrating MKBE is able to utilize different multimodal values for modeling numerical information. Further, the neural decoder performs better than the search-based one, showing the importance of proper decoder, even for finite, enumerable sets.
Along the same line, Table~\ref{fig:gm} shows genre prediction accuracy on $10\%$ of held-out MovieLens dataset. 
Again, the model that uses all the information outperforms other methods.

\begin{table}[tb] 
     
      \centering

        \small
        \caption{\textbf{Evaluating Generated Titles} for MovieLens using movies embeddings conditioned on just the ratings (R) and all the information. We present the accuracy of the users in guessing whether the generated title for a movie was real (yes/no), and genre of the movie ($4$ choices).}

        \label{tab:tr}
            \begin{tabular}{lcc}
            \toprule
            \bf Models & \bf Real vs Fake &\bf Genre\\
            \midrule
            R  &63&27.2\\
            R+M+U+T+P &73&41.6\\ 
            Reference &90&68\\
            \bottomrule
            \\
            \end{tabular}
            \postspace{}
\end{table}
\begin{table}[tb] 
      \centering
        \small
        \caption{\textbf{Evaluating Generated Text and Images} for YAGO using entity embeddings conditioned on just the links (S) or all information. We present the accuracy of the users in guessing whether the generated text/image for a person was real (yes/no), gender of the person, age ($<$35, or $\geq$35), and occupation ($3$ choices).}
        \label{tab:gen_numbers}
            \begin{tabular}{clcccc}
            \toprule
            &\bf Models & \bf Real &\bf Gender&\bf Age&\bf Occup.\\
            \midrule
            \parbox[t]{2mm}{\multirow{3}{*}{\rotatebox[origin=c]{90}{descrip.}}}&S &57.1&72.1&59&71.4\\ 
            &S+N+D+I &59.2&77.2&63.4&78.6\\ 
            &Reference &67.8&83.2&69.5&90.4\\
            \midrule
            \parbox[t]{2mm}{\multirow{3}{*}{\rotatebox[origin=c]{90}{images}}}&S &60&67&53&43\\ 
            &S+N+D+I & 67&77&53&52\\
            &Reference &96&1.0&83&82\\
            \bottomrule
            \\
            \end{tabular}
        \postspace{}
        \postspace{}

\end{table}
     
     


\begin{table}[tb]
    \centering
    \small
    \caption{\textbf{Generated Descriptions} for "Carles Puyol" (and the corresponding reference from the DBpedia) by embeddings trained from just the links (S) and all of the information (S+N+D+I).}
    \label{tab:ex}
          	        \setlength{\tabcolsep}{3pt}
    \begin{tabular}{lp{60mm}}
    \toprule
    \bf Model & \bf Generated Descriptions \\
    \midrule
   Reference & $\langle$subject$\rangle$ (born 13 April 1978) is a Spanish retired professional footballer.\\
    \addlinespace
    Only S & $\langle$subject$\rangle$ (born 25 January 1949) is a Georgian football coach and former professional player.\\
    \addlinespace
    S+N+D+I & $\langle$subject$\rangle$ (born 22 April 1967) is an English former football player.
    \\
    \bottomrule
    \end{tabular}
        \postspace{}
\end{table}




  

\begin{table}[tb]
    \centering
    \small
    \caption{\textbf{Generated Images} for YAGO. We consider athletes, and male and female celebrities, and compare their reference images with corresponding ones generated from all the information.} 
    \label{tab:ex-im}
     	        \setlength{\tabcolsep}{1pt}
    \begin{tabular}{llcc}
    \toprule
    & &\bf Reference & \bf S+N+D+I \\
  \cmidrule(lr){3-3}  \cmidrule(lr){4-4}

     &\multirow{-7}{*}{\rotatebox[origin=c]{90}{\textbf{Athletes}}}&\includegraphics[width=.2\textwidth]{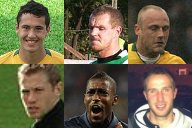} &
     \includegraphics[width=.2\textwidth]{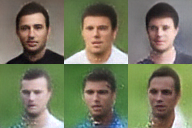} \\
     
      \multirow{-7}{*}{\rotatebox[origin=c]{90}{\textbf{Male}}}
      &\multirow{-7}{*}{\rotatebox[origin=c]{90}{\textbf{celebrities}}} &\includegraphics[width=.2\textwidth]{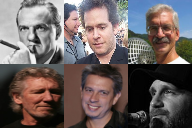} &
     \includegraphics[width=.2\textwidth]{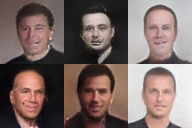} \\

      \multirow{-7}{*}{\rotatebox[origin=c]{90}{\textbf{Female}}}
     &\multirow{-7}{*}{\rotatebox[origin=c]{90}{\textbf{celebrities}}} &\includegraphics[width=.2\textwidth]{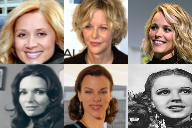} &
     \includegraphics[width=.2\textwidth]{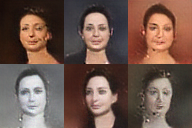} \\



   \bottomrule
  
    \end{tabular}
    \postspace{}
\end{table}
\para{MovieLens Titles}
 For generating movie titles, we randomly consider $200$ of them as test, $100$ as validation, and the remaining ones as training data. The goal here is to generate titles for movies in the test data using the previously mentioned GAN structure. To evaluate our results we conduct a human experiment on Amazon Mechanical Turk (AMT) asking participant two questions: (1)~whether they find the movie title real, and (2)~which of the four genres is most appropriate for the given title. We consider $30$ movies each as reference titles, fake titles generated from only ratings as conditional data, and fake titles conditioned on all the information. Further, each question was asked for $3$ participants, and the results computed over the majority choice are shown in Table~\ref{tab:tr}. 
Fake titles generated with all the information are more similar to reference movie titles, demonstrating that the embeddings that have access to more information effectively generate higher-quality titles. 


\para{YAGO Descriptions} 
The goal here is to generate descriptive text for entities from their embeddings.
Since the original descriptions can be quite long, we consider first sentences that are less than 30 tokens, resulting in $96,405$ sentences. 
We randomly consider $3000$ of them as test, $3000$ as validation, and the remaining as training data for the decoder. 
To evaluate the quality of the generated descriptions, and whether they are appropriate for the entity, we conduct a user study asking participants if they can guess the realness of sentences and the occupation (\emph{entertainer}, \emph{sportsman}, or \emph{politician}), gender, and age (above or below $35$) of the subject entity from the description.
We provide $30$ examples for each model asking each question from $3$ participants and calculate the accuracy of the majority vote. 
The results presented in Table~\ref{tab:gen_numbers} show that the models are fairly competent in informing the users of the entity information, and further, 
descriptions generated from embeddings that had access to more information outperforms the model with only structured data. 
Examples of generated descriptions are provided in Table~\ref{tab:ex}
 (in addition to screenshots of user study, more examples of generated descriptions, and MovieLens titles are provided in supplementary materials). 

\para{YAGO Images}
Here, we evaluate the quality of images generated from entity embeddings by humans ($31,520$, split into train/text). Similar to descriptions, we conduct a study asking users to guess the realness of images and the occupation, gender, and age of the subject. 
We provide $30$ examples for each model asking each question from $3$ participants, and use the majority choice.

The results in Table~\ref{tab:gen_numbers} indicate that the images generated with embeddings based on all the information are more accurate for gender and occupation. 
Guessing age from the images is difficult since the image on DBpedia may not correspond to the age of the person, i.e. some of the older celebrities had photos from their youth. 
Examples of generated images are shown in Table~\ref{tab:ex-im}.

\section{Discussion and Limitations}
An important concern regarding KB embedding approaches is their scalability. While large KBs are a problem for all embedding-based link prediction techniques, MKBE is not significantly worse than existing ones because we treat multimodal information as additional triples. 
Specifically, although multimodal encoders/decoders are more expensive to train than existing relational models, the cost is still additive as we are effectively increasing the size of the training dataset. 

In addition to scalability, there are few other challenges when working with multimodal attributes. 
Although multimodal evidence provides more information, it is not at all obvious which parts of this additional data are informative for predicting the relational structure of the KB, and the models are prone to overfitting. 
MKBE builds upon the design of neural encoders and decoders that have been effective for specific modalities, and the results demonstrate that it is able to utilize the information effectively. 
However, there is still a need to further study models that capture multimodal attributes in a more efficient and accurate manner. 

Since our imputing multimodal attributes model is based on GAN structure and the embeddings learned from KB representation, the generated attributes are directly limited by the power of GAN models and the amount of information in the embedding vectors. 
Although our generated attributes convey several aspects of corresponding entities, their quality is far from ideal due to the size of our datasets (both of our image and text datasets are order of magnitude smaller than common datasets in the existing text/image genration literature) and the amount of information captured by embedding vectors (the knowledge graphs are sparse). 
In future, we would like to (1)~expand multimodal datasets to have more attributes (use many more entities from YAGO), and (2)~instead of using learned embeddings to generate missing attributes, utilize the knowledge graph directly for generation.


\section{Conclusion}
Motivated by the need to utilize multiple sources of information, such as text and images, to achieve more accurate link prediction, we present a novel neural approach to multimodal relational learning. 
We introduce MKBE, a link prediction model that consists of (1)~a compositional encoding component to jointly learn the entity and multimodal embeddings to encode the information available for each entity, and (2)~adversarially trained decoding component that use these entity embeddings to impute missing multimodal values.
We enrich two existing datasets, YAGO-10 and MovieLens-100k, with multimodal information to introduce benchmarks. 
We show that MKBE, in comparison to existing link predictors DistMult and ConvE, can achieve higher accuracy on link prediction by utilizing the multimodal evidence. 
Further, we show that MKBE effectively incorporates relational information to generate high-quality multimodal attributes like images and text. 
We have release the datasets and the open-source implementation of our models at \url{https://github.com/pouyapez/mkbe}. 



\section*{Acknowledgements}
We would like to thank Zhengli Zhao, Robert L. Logan IV, Dheeru Dua, Casey Graff, and the anonymous reviewers for their detailed feedback and suggestions.  This work is supported in part by Allen Institute for Artificial Intelligence (AI2) and in part by NSF award \#IIS-1817183. The views expressed are those of the authors and do not reflect the official policy or position of the funding agencies.

\bibliography{pouya}
\bibliographystyle{acl_natbib}

\end{document}

%% file: macros.tex
\newcommand{\R}{\ensuremath{\mathcal{R}}}
\newcommand{\M}{\ensuremath{\mathcal{M}}}
\newcommand{\Real}{\mathbb{R}}

\newcommand{\cut}[1]{}